\documentclass[letterpaper, 10 pt, conference]{ieeeconf}  

\IEEEoverridecommandlockouts                              

\usepackage{cite}
\usepackage{amsmath,amssymb,amsfonts}
\usepackage{algorithmicx}
\usepackage{algorithm}
\usepackage{algpseudocode}
\algrenewcommand\algorithmicrequire{\textbf{Input:}}
\algrenewcommand\algorithmicensure{\textbf{Output:}}
\usepackage{graphicx}
\usepackage{textcomp}
\usepackage{xcolor}
\usepackage{color} 

\newcommand{\rev}[1]{\textcolor{black}{#1}}
\newcommand{\revi}[1]{\textcolor{black}{#1}}
\usepackage{pgfplots}
\newcommand\Rb{\mathbb{R}}
\usepackage[font=small]{caption}

\usepackage{csquotes}
\usepackage{tikz}
\usepackage{pgfplots}
\usepackage{xcolor}
\usepackage{subcaption}
\pgfplotsset{compat=1.18}
\usetikzlibrary{positioning}
\usetikzlibrary{decorations.pathreplacing}
\usetikzlibrary{calc}
\usetikzlibrary{positioning,shapes,shadows,arrows}
\usepackage{todonotes}
\usepackage{multirow}
\usepackage{leftidx}

\pgfplotsset{
every axis/.append style={
  axis line style={->}, 
  legend style={font=\scriptsize},
  label style={font=\scriptsize},
  title style={font=\scriptsize},
  tick label style={font=\scriptsize},
  axis x line*=bottom,
  axis y line*=left,
  }
}

\makeatletter
\let\NAT@parse\undefined
\makeatother
\usepackage{hyperref}

\title{\LARGE \bf
UNO Push: Unified Nonprehensile Object Pushing via\\Non-Parametric Estimation and Model Predictive Control}

\author{Gaotian Wang, Kejia Ren, and Kaiyu Hang
\thanks{The authors are with the Department of Computer Science, Rice University, Houston, TX 77005, USA. {\{\tt\small gwang, kr43, kaiyu.hang\}@rice.edu.} This work is supported by NSF grant FRR-2133110 and Rice University Funds.}
}

\begin{document}
\maketitle

\begin{abstract}



Nonprehensile manipulation through precise pushing is an essential skill that has been commonly challenged by perception and physical uncertainties, such as those associated with contacts, object geometries, and physical properties. 
For this, we propose a unified framework that jointly addresses system modeling, action generation, and control. 
While most existing approaches either heavily rely on \emph{a priori} system information for analytic modeling, or leverage a large dataset to learn dynamic models, our framework approximates a system transition function via non-parametric learning only using a small number of exploratory actions (\emph{ca.} $10$). 
The approximated function is then integrated with model predictive control to provide precise pushing manipulation. 
Furthermore, we show that the approximated system transition functions can be robustly transferred across novel objects while being online updated to continuously improve the manipulation accuracy. 
Through extensive experiments on a real robot platform with a set of novel objects and comparing against a state-of-the-art baseline, we show that the proposed unified framework is a light-weight and highly effective approach to enable precise pushing manipulation all by itself.
Our evaluation results illustrate that the system can robustly ensure millimeter-level precision and can straightforwardly work on any novel object.

\end{abstract}
\section{Introduction}

\begin{figure}[t]
    \centering
    
    \includegraphics[width=1\columnwidth]{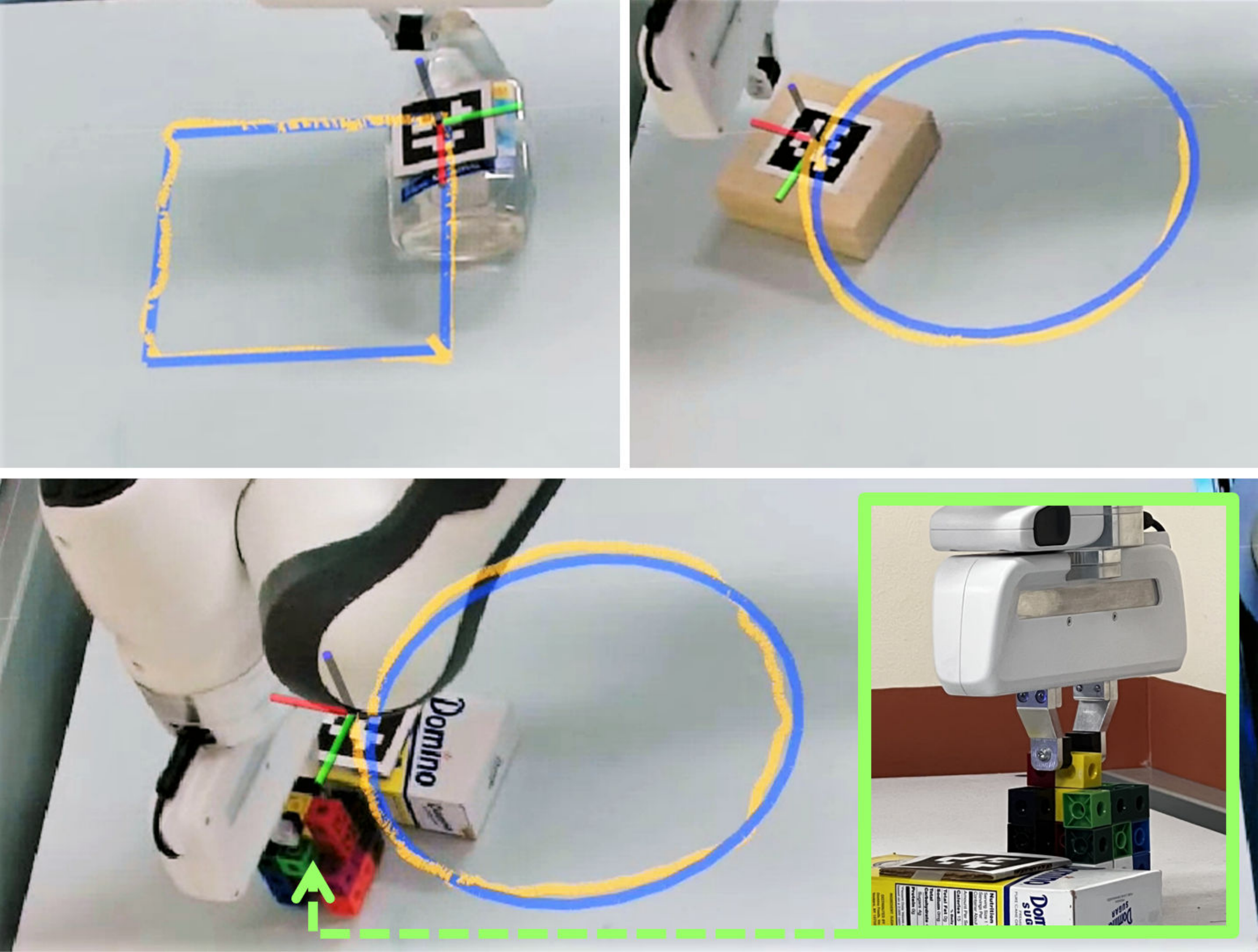}
    
    \caption{A robot manipulator is tasked to manipulate an unknown object to trace reference trajectories (blue). Without analytically modeling the contacts or other physical components of the system, our UNO Push framework enables precise manipulation (yellow trajectories) by pushing the object with the gripper of the robot (top), or by an unknown object grasped by the gripper (bottom). }
    \label{fig:first}
    \vspace{-0.5cm}
\end{figure}

Nonprehensile manipulation actions such as pushing, sliding, and toppling \cite{lynch1996stable, Hang2019, Lynch99}, can provide a rich set of physical possibilities for robots to interact with objects. More commonly than other motion primitives, pushing has been widely employed as a key component in manipulation systems to handle tasks where grasping is unnecessary or infeasible. In general, pushing-based manipulation is formulated either as a large-scale multi-objects rearrangement problem \cite{song2020multi, huang19, haustein2015kinodynamic, agboh2018real}, or as a problem concerned with the precise motion control between a robot and a pushed object \cite{zhou2019pushing}. While both formulations are challenged by the intricate dynamics and various uncertainties in perception and physics, this work focuses on the precise control of pushing manipulation, as exemplified in Fig.~\ref{fig:first} with the goal of minimizing the requirements on sensing and prior knowledge while optimizing the manipulation precision.

Analytic methods for precise pushing are traditionally quite heavy in terms of the involved system components, including contact analysis, object shape representations, modeling of system transitions, physical uncertainties, action generation, and control \cite{cheng21, mason1999progress, koval2016pre, zhou2019pushing, hogan2020reactive, Bertoncelli20}. Nevertheless, such complex compositions often make the system integration prohibitively complex and render the solutions not generalizable nor scalable to different task setups, as limited by many modeling simplification assumptions.
Alternatively, data-driven approaches \cite{yuan2019end, bauza2019omnipush, kloss2022combining, bauza2017probabilistic, ajay2018augmenting, yu2016more, bauza2018data} have shown unprecedented capabilities in handling complex manipulation tasks. Although data can enable such methods, it is also a limiting factor when the task setup changes or the perception design varies across systems, which would normally require the entire system to be trained again with a large amount of new data.

To address the aforementioned challenges, this work builds our system upon two insights. First, an approximated inaccurate system model can be used to close the control loop and ensure high precision. Second, this approximated model, which presents certain discrepancies against the real physical system, can still offer good performance when
adaptively updated online to match the observed physical outcomes.


To this end, this work proposes a \emph{unified} framework that addresses system modeling, action generation, and control of precise pushing all through non-parametric estimation, named as UNO Push for \emph{Unified Nonprehensile Object Pushing}.
The framework first builds the system transition model via \emph{light-weight} non-parametric estimation to directly map from the robot actions to the object motions.
Without requiring any \emph{a priori} knowledge about the object or robot contact geometries or physics, nor any large dataset or \revi{object-specific} offline training, our method can effectively approximate a transition model using a few exploratory actions (\emph{ca.} $10$). 
Then, via closed-loop Model Predictive Control built upon the approximated model, our framework generates real-time actions by observing the system state.
Real-time feedback is used to adaptively update the approximated model online to continuously improve the control performance.

Through extensive experiments, we show that the \emph{unified light-weight} UNO Push can directly work on novel objects using highly approximated models, which can be easily transferred and adapted to precisely manipulate other objects. 
By comparing against a state-of-the-art baseline approach, we show that our light-weight framework can ensure high manipulation precision without sophisticated modeling or \revi{object-specific pre-training}.
Through experiments of pushing by the robot gripper and by a grasped unmodeled object (Fig.~\ref{fig:first}), we illustrate the possibility of deploying our method on different robots without requiring any remodeling. \rev{We further show that even under unknown external perturbations, such as pushing through a cluttered area (Fig.~\ref{fig:obstacle}), our UNO Push is able to effectively handle the uncertainties in the environment to ensure precise pushing.}

\rev{As will be discussed in detail, the unified framework of data efficient model approximation, online model update, and closed-loop control, provides a very low barrier for UNO Push to be flexibly employed in real-world tasks without much pre-requisites. This is especially useful when the contact geometries and physics are unknown or when the task conditions do not allow for pre-training on the target object, making analytical modeling or offline training infeasible. The key contributions of our work are:}
\rev{\begin{itemize}
    \item A unified framework that addresses system modeling, action generation, and control of precise pushing all through non-parametric estimation;
    \item System motion models built through a small number of exploratory actions;
    \item Precise pushing manipulation with imprecisely approximated system models, which are continuously updated online using in-task experiences.
\end{itemize}}

\vspace{-10pt}
\rev{\section{Related Work}}

\emph{Analytic Models:} Precise control of planar pushing can be achieved through analytically modeling the control laws. Assuming necessary \revi{\textit{a priori} }knowledge are available, e.g., friction coefficient, object mass distribution, and object geometry, \rev{differentiable} system transition models can be derived under some common simplification assumptions, such as point contact and quasi-static physics \cite{zhou2019pushing}. Thereafter, action generation can be enabled through various optimization formulations \cite{hogan2020reactive}, and control loops are normally closed by nonlinear control schemes, such as model predictive control \cite{Bertoncelli20}. However, as existing analytic approaches aim at building models that are unnecessarily accurate, they are commonly not generalizable or scalable, especially against unknown and uncertain task setups. 

\emph{Simulation-based Planning:} For large-scale rearrangement manipulation problems, pushing actions are planned to sequentially reconfigure the system states \cite{haustein2015kinodynamic}. In such problems, algorithms such as kinodynamic motion planning and trajectory optimization are more concerned with the discrete transitions between states, while the precise motions along the transitions are often ignored. Although also challenged by physical uncertainties \cite{agboh2018real}, simulation-based planning methods are fundamentally different from precise control methods, such as the one addressed in this work, as they do not need to build the system models.


\emph{Data-Driven Approaches:} Meta-learning of dynamic models \cite{bauza2019omnipush}, composite analytic and learned models \cite{kloss2022combining}, probabilistic approaches \cite{bauza2017probabilistic, bauza2018data}, stochastic neural networks \cite{ajay2018augmenting}, and large public datasets \cite{yu2016more}, together with many other data-driven approaches, have shown unprecedented capabilities in handling complex tasks. \rev{Being the state-of-the-art work in precise pushing control, the study in \cite{bauza2017probabilistic, bauza2018data} have demonstrated that effective control performance can be achieved when system motion models are pre-trained on objects of uniform mass distributions, with additional assumptions that the shape of the object is known or well approximated.} However, similar to data-driven approaches in other problems, data is a major limitation for model generalization, and a small change in task setup or perception design can often require a new dataset to be collected for transferring the model. In contrast, this work shows that an inaccurate model can be approximated with a small amount of data and a light-weight model, and can be integrated into a unified framework, while being updated online, \rev{to enable precise pushing control of objects of unknown geometries and physical properties.}

\vspace{15pt}
\section{Problem Formulation}
\label{sec:prob}

In this work, we are interested in the problem of pushing-based nonprehensile manipulation, where a robot manipulator is tasked to continuously push a single object to trace some desired trajectories of the object.
We assume the motion of the object to be planar sliding without rolling or flipping, in a quasi-static manner.
As such, the controlled object's motion can be modeled by a discrete-time dynamical system.
We denote the configuration of the object at time $t$ by $X_t \in SE(2)$, and define $\mathcal{U}$ to be the set consisting of all the allowed controls that the robot can execute to push the object. 
The object's configuration evolves according to the following deterministic system dynamics:
\begin{equation}
    X_{t+1} = f(X_t, u_t)
\label{eq:f}
\end{equation}
where $f: SE(2) \times \mathcal{U} \mapsto SE(2)$ is the system transition function and $u_t \in \mathcal{U}$ is the control executed by the robot.

\subsection{Manipulation Model Representation}
\label{sec:rep}

\emph{Object's Configuration:}
We use a homogeneous representation for the object's configuration. That is, $X_t \in SE(2) $ is represented by a $3 \times 3$ transformation matrix.
Given the object's configurations at adjacent time steps, $X_t$ and $X_{t+1}$, the rigid body motion of the object at time $t$ is defined by $\leftidx{^b}{g}_t = X_t^{-1} \cdot X_{t+1} \in SE(2)$, where $X_t^{-1}$ is the inverse of $X_t$.
In this definition, the object's rigid body motion $\leftidx{^b}{g}_t$ is always specified in the object's body frame.
(\emph{Note:} Throughout the paper, a left superscript $b$ in the notation indicates that the variable is defined in the object's body frame; otherwise, it is defined in the spatial frame.)

Furthermore, we define a distance function in the object's configuration space by $\Delta: SE(2) \times SE(2) \mapsto \Rb$.
We calculate the distance between two arbitrary configurations of the object by the \revi{weighted} summation of the difference in their orientations and the Euclidean distance between their positions.

\begin{figure}[h]
\vspace{-5pt}
\centering
\includegraphics[width=\columnwidth]{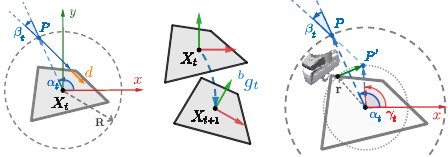}
\caption{\textit{Left}: The representation of the control, through two angles $\alpha_t$ and $\beta_t$ in the object's body frame; \textit{Middle}: The rigid body motion of the object $\leftidx{^b}{g}_t$; \textit{Right}: The smoothened execution strategy of control. Instead of retreating the gripper back to the point $P$, the robot moves the gripper to a point $P'$ closer to the object.}
\label{fig:re}
\vspace{-10pt}
\end{figure}

\emph{Control:}
\rev{Typically, most control models for object pushing require precise locations of the robot-object contacts to be available. However, this is often impractical in the real-world tasks when handling objects of unknown shapes.} To enable effective pushing actions for objects of arbitrary shapes, we represent the control by two angles defined in the object's body frame: $\leftidx{^b}{u}_t = (\alpha_t, \beta_t)$.
As illustrated in Fig.~\ref{fig:re}, we virtually attach a circle to the object's body frame, whose radius $R$ is set larger than the size of the object.
To execute a control, the robot first moves its end-effector to a point $P$ on the circle, determined by the angle $\alpha_t$.
Then, the robot continuously moves the end-effector towards the object in the direction determined by an offset angle $\beta_t$ until it has traveled a constant distance $d$ \rev{after detecting the object's motion}. As such, the controls are specified relative to the object's body frame.

\emph{System Transition Models:}
Since the control is defined in the object's body frame,
for the same object, its rigid body motion is independent of its configuration $X_t$.
Therefore, we can represent the motion of the object by a transition function $\Gamma: \mathcal{U} \mapsto SE(2)$ invariant to the object's configuration, which maps a control to the object's rigid body motion.
The system transition model in Eq.~\eqref{eq:f} can be re-written as:
\begin{equation}
    X_{t+1} = X_t \cdot \leftidx{^b}{g}_t = X_t \cdot \Gamma(\leftidx{^b}{u}_t)
\end{equation}

In addition, we need an inverse model of the system transitions, $\Gamma^{-1}: SE(2) \mapsto \mathcal{U}$, which infers the control to be executed given a desired rigid body motion of the object.

\begin{figure}[h]
\vspace{-10pt}
\begin{algorithm}[H]
    \caption{Precise Pushing via UNO Push}
    \label{alg:uno}
    \footnotesize
    \begin{algorithmic}[1]
    \Require Object's initial configuration $X_0$, reference trajectory $\mathcal{Y}$,
    a boolean argument $s$ indicating whether to \rev{learn} models from scratch,
    a distance threshold $\delta$
            \State $\Gamma, \Gamma^{-1} \gets \Call{LearnModels}{X_0, s}$ \hfill \Comment{Alg.~\ref{alg:model_init}}
        \State $t \gets 0$
            \While{$\Delta(X_t, Y_M) > \delta$} \hfill \Comment{Last Waypoint $Y_M$ Not Reached}
                \State $\leftidx{^b}{u}_t \gets \Call{MPC}{X_t, \mathcal{Y}}$ \hfill \Comment{Alg.~\ref{alg:mpc}}
                \State $X_{t+1} \gets \Call{SmoothenedExecute}{X_t, \leftidx{^b}{u}_t}$ \hfill\Comment{Alg.~\ref{alg:exec}}
                \State $\Gamma, \Gamma^{-1} \gets \Call{UpdateModels}{\leftidx{^b}{u}_t, X_t, X_{t+1}}$ \hfill \Comment{Alg.~\ref{alg:model_update}}
                \State $t \gets t+1$
            \EndWhile
    \end{algorithmic}
\end{algorithm}
\vspace{-20pt}
\end{figure}

\subsection{Precise Pushing Problem}
Starting with the initial configuration of the object $X_0 \in SE(2)$, the robot is required to find and execute a sequence of controls, as defined in Sec.~\ref{sec:rep}, to push the object to $M$ desired configurations sequentially.
These configurations compose a reference trajectory, represented by a sequence $\mathcal{Y} = \{Y_1, \cdots, Y_M\}$ where $Y_1, \cdots, Y_M \in SE(2)$.

In general, generating actions to accomplish the aforementioned manipulation task requires an accurate model of the system dynamics.
However, analytical system models $\Gamma$ and $\Gamma^{-1}$ are not feasible as they require accurate object geometries and physical parameters such as the friction coefficient, which are not available without ideal and sophisticated sensing capability.
Moreover, analytical $\Gamma$ and $\Gamma^{-1}$ are difficult to generalize on different objects, limiting their applications in the real world.
To this end, we propose to approximate the system models by using manipulation experiences observed online,
and integrate the approximated models into a Model Predictive Control (MPC) framework for generating effective actions.
In this way, we unify the model approximation, action generation, and control into a light-weight yet efficient framework.
The proposed framework, UNO Push, is presented in Alg.~\ref{alg:uno}.

\section{Non-parametric Model Estimation}
\label{sec:model}
To approximate the system models $\Gamma$ and $\Gamma^{-1}$ defined in Sec.~\ref{sec:rep} without requiring a large amount of data or prior information about the system, we propose to represent $\Gamma$ and $\Gamma^{-1}$ by non-parametric models and estimate them through Gaussian Process Regression (GPR).
As we try to keep our framework light-weight and limit the amount of data to be very small,
the models $\Gamma$ and $\Gamma^{-1}$ can be estimated and updated online while the robot is manipulating the object. 
The predictions made by the approximated models will then be used to generate real-time actions.

\begin{figure}[htbp]
\centering
\includegraphics[width=\columnwidth]{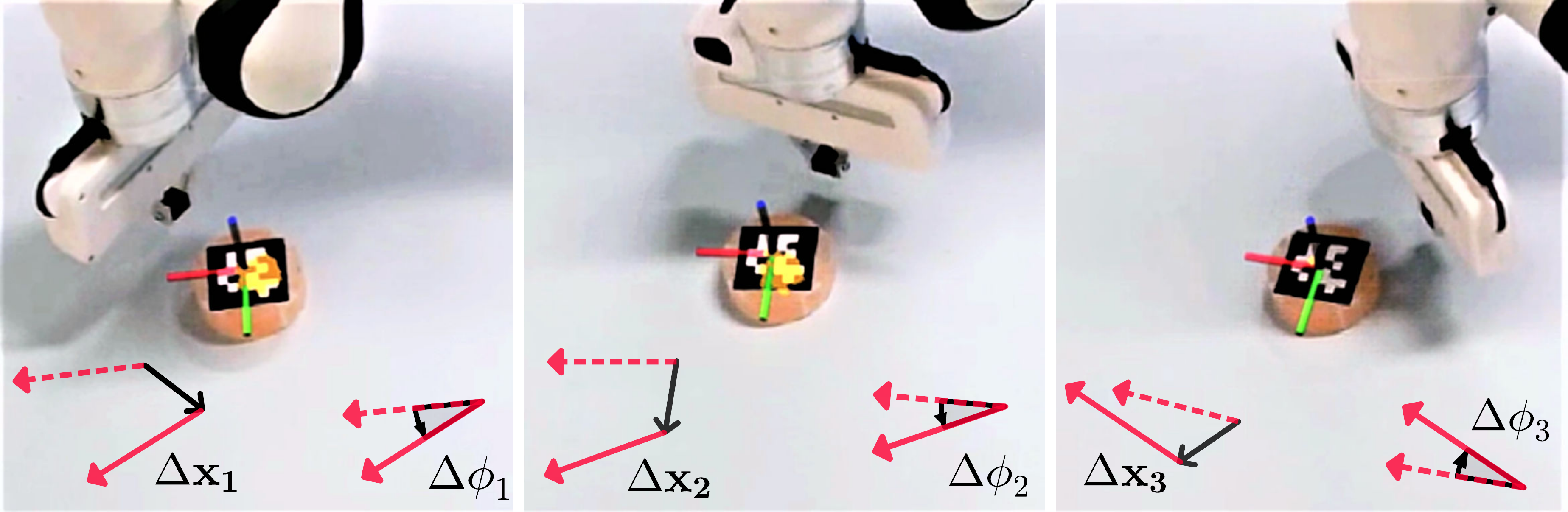}
\caption{Three example data points collected by pushing a cylinder through random controls. The red dashed and solid arrows represent the x-axis of the object's body frame before and after the push, respectively. The object's configuration has been changed through translation $\Delta \mathbf{x}$ and rotation $\Delta\phi$.
\rev{In our experiments (Sec.~\ref{sec:experiments}), we applied the model learned on the cylinder object to directly manipulate other objects.}}
\label{fig:sample}
\vspace{-10pt}
\end{figure}

\subsection{Model \rev{Learning} }
\label{sec:model_init}

To \rev{learn} the approximated models of $\Gamma$ and $\Gamma^{-1}$,
we build a training dataset $\mathcal{D} = \{(\leftidx{^b}{\hat{u}}_i, \leftidx{^b}{\hat{g}}_i)\}_{i=0}^{N-1}$ of $N$ data points through real-world manipulation.
Each data point in $\mathcal{D}$ is a pair of control $\leftidx{^b}{\hat{u}}_i \in \mathcal{U}$ the robot has executed and the observed rigid body motion of the object $\leftidx{^b}{\hat{g}}_i \in SE(2)$.
As detailed in Alg.~\ref{alg:model_init} and illustrated in Fig.~\ref{fig:sample}, we uniformly sample $N$ controls within their allowed ranges and execute each sampled control on the robot to manipulate the object by pushing.
In practice, the two angles of the control are sampled from $\alpha \in [0, 2\pi]$ radians and $\beta \in [-0.2, 0.2]$ radians.
The range of $\beta$ was made narrow to increase the probability of the robot making contact with the object.
Meanwhile, the object's motion $\leftidx{^b}{\hat{g}}_i$ resulting from the execution of $\leftidx{^b}{\hat{u}}_i$ is observed by sensors.
In the end, we add all these sampled controls and their corresponding observations into the training dataset $\mathcal{D}$ and regress the initial models of $\Gamma$ and $\Gamma^{-1}$ via GPR, to learn the underlying relationship between the controls and the object's motions.
Both models $\Gamma$ and $\Gamma^{-1}$ are regressed with the same dataset $\mathcal{D}$, by swapping the domain and codomain of the data.
Although trained with the same dataset, the estimated $\Gamma$ and $\Gamma^{-1}$ are not constrained to form a closed loop, that is, $\Gamma^{-1}(\Gamma (u)) \neq u$.

\begin{figure}[t]
\vspace{-5pt}
\begin{algorithm}[H]
    \caption{LearnModels($\cdot$)}
    \label{alg:model_init}
    \footnotesize
    \begin{algorithmic}[1]
    \Require Object's initial configuration $X_0$, a boolean argument $s$ indicating whether to \rev{learn} models from scratch
    \Ensure Learned models $\Gamma$ and $\Gamma^{-1}$
        \If{$s ==$ True} \hfill \Comment{\rev{Learn} Models from Scratch}
        \State $\mathcal{D} \gets \{\}$ \hfill \Comment{Training Dataset}
        \For{$i=0, \cdots, N-1$}
            \State $\alpha \gets \Call{Uniform}{0, 2\pi}$ \hfill \Comment{Uniform Sampling}
            \State $\beta \gets \Call{Uniform}{-0.2, 0.2}$
            \State $\leftidx{^b}{\hat{u}}_i \gets (\alpha, \beta)$ \hfill \Comment{Random Control}
            \State $X_{i+1} \gets \Call{Execute}{\leftidx{^b}{\hat{u}}_i}$ \hfill \Comment{Observe Object's Configuration}
            \State $\leftidx{^b}{\hat{g}}_i \gets X_{i}^{-1} \cdot X_{i+1}$ \hfill \Comment{Object's Rigid Body Motion}
            \State $\mathcal{D} \gets \mathcal{D}\cup \{(\leftidx{^b}{\hat{u}}_i, \leftidx{^b}{\hat{g}}_i)\}$
        \EndFor
    \State $\Gamma, \Gamma^{-1} \gets \Call{GPR}{\mathcal{D}}$ \hfill \Comment{Gaussian Process Regression}
    \Else
        \State $\Gamma, \Gamma^{-1} \gets \Call{CopyFromOld}$ \hfill \Comment{Copy From Previous Tasks}
    \EndIf
    \State \Return $\Gamma, \Gamma^{-1}$ 
    \end{algorithmic}
\end{algorithm}
\vspace{-30pt}
\end{figure}

Based on the intuition that motion models of different objects have similarities in their patterns, our framework has the option to \rev{transfer the models learned from previous manipulation tasks as the manipulation models for a novel object}.
In such cases, when manipulating a novel object, our framework can skip the real-world data collection step (lines 2-11 of Alg.~\ref{alg:model_init}) and directly use the models \rev{transferred} from previous tasks of manipulating a different object, to speed up the current manipulation task.
Even though the models transferred from manipulating a different object are not accurate enough at the beginning, they can be updated online to improve the manipulation performance.

\subsection{Online Model Update}
\label{sec:model_update}

When the robot uses the \rev{learned} models of $\Gamma$ and $\Gamma^{-1}$ in Sec.~\ref{sec:model_init} to push the object, it at the same time keeps exploring the system transition models.
Hence, we propose to adaptively update the model online using the newly executed actions as detailed in Alg.~\ref{alg:model_update}. Specifically, whenever a control $\leftidx{^b}{u}_t$ is executed to manipulate the object, $\leftidx{^b}{u}_t$ associated with the observed object's motion $\leftidx{^b}{g}_t$ will be added to the training dataset $\mathcal{D}$ to update the models $\Gamma$ and $\Gamma^{-1}$.

\begin{figure}[t]
\begin{algorithm}[H]
    \caption{UpdateModels($\cdot$)}
    \label{alg:model_update}
    \footnotesize
    \begin{algorithmic}[1]
    \Require Last executed control $\leftidx{^b}{u}_t$, object's configuration before last execution $X_t$, object's configuration after last execution $X_{t+1}$
    \Ensure Updated Models $\Gamma$ and $\Gamma^{-1}$
    \State $\mathcal{D} \gets \Call{AcquireDataset()}{}$ \hfill \Comment{Current Training Dataset}
    \For{$(\leftidx{^b}{\hat{u}}_i, \leftidx{^b}{\hat{g}}_i) \in \mathcal{D}, i=1, \cdots, \lvert \mathcal{D} \rvert$}
        \If{$\lVert \leftidx{^b}{u}_t - \leftidx{^b}{\hat{u}}_i \rVert < \epsilon$}
            \State $\mathcal{D}.\Call{Remove}{(\leftidx{^b}{\hat{u}}_i, \leftidx{^b}{\hat{g}}_i)}$
        \EndIf
    \EndFor
    \State $\leftidx{^b}{g}_t \gets X_t^{-1} \cdot X_{t+1}$ \hfill \Comment{Object's Rigid Body Motion}
    \State $\mathcal{D} \gets \mathcal{D} \cup \{(\leftidx{^b}{u}_t, \leftidx{^b}{g}_t)\}$
    \State $\Gamma, \Gamma^{-1} \gets \Call{GPR}{\mathcal{D}}$ \hfill \Comment{Gaussian Process Regression}
    \State \Return $\Gamma, \Gamma^{-1}$ 
    \end{algorithmic}
\end{algorithm}
\vspace{-30pt}
\end{figure}

Moreover, if the newly executed control $\leftidx{^b}{u}_t$ is very similar to any control $\leftidx{^b}{\hat{u}}_i$ which already exists in the dataset $\mathcal{D}$ (i.e., their difference $\lVert \leftidx{^b}{u}_t - \leftidx{^b}{\hat{u}}_i \rVert$ is less than a threshold $\epsilon$), the outdated control $\leftidx{^b}{\hat{u}}_i$ will be removed from the dataset $\mathcal{D}$.
This mechanism helps keep the size of the dataset small to facilitate efficient model approximation.
More importantly, it enforces updating the models with the most recent data points, which are more relevant to the current task setup and the physical state of the robot-object system.

Importantly, online model update is useful especially when transferring the approximated models between manipulations of different objects.
When the models are transferred by previous experiences of manipulating an old object, updating the models with online data can ensure fast adaptation of the models to the manipulation of the new object.

\section{MPC-based Action Generation}
\label{sec:mpc}

With the system models $\Gamma$ and $\Gamma^{-1}$ approximated and adaptively updated online in Sec.~\ref{sec:model},
a model-based control scheme can be applied to generate real-time controls taking into account the predictions made by 
$\Gamma$ and $\Gamma^{-1}$.
To this end, as illustrated in Alg.~\ref{alg:mpc}, we integrate the approximated models $\Gamma$ and $\Gamma^{-1}$ in a Model Predictive Control (MPC) framework to close the control loop, for generating effective robot actions to precisely manipulate the object.

\begin{figure}

\begin{algorithm}[H]
    \caption{Model Predictive Control (MPC)}
    \label{alg:mpc}
    \footnotesize
    \begin{algorithmic}[1]
    \Require Observed object's configuration $X_t$, reference trajectory $\mathcal{Y}$
    \Ensure Generated control for execution $\leftidx{^b}{u}_t$
    \For{$q = 1, \cdots, Q$}
        \State $\mathcal{T}^q \gets \{ \hat{X}_0^q = X_t\}$ \hfill \Comment{Simulated Trajectory}
        \For{$k = 0, \cdots, L-1$}
            \State $j^* \gets \arg\min_j \Delta (\hat{X}_k^q, Y_j)$ \hfill \Comment{Nearest Waypoint in $\mathcal{Y}$}
            \State $\leftidx{^b}{\hat{g}}_k^q \gets \hat{X^q_k}^{-1} \cdot Y_{j^*+1}$ \hfill \Comment{Desired Motion of Object}
            \State $g_\xi \gets \Call{RandTransformation()}{}$ \hfill \Comment{Perturbation}
            \State $\leftidx{^b}{\hat{u}}_k^q = \Gamma^{-1} (\leftidx{^b}{\hat{g}}_k^q \cdot g_\xi)$ \hfill \Comment{Predicted Control}
            \State $\hat{X}_{k+1}^q \gets \hat{X}_k^q \cdot \Gamma(\leftidx{^b}{\hat{u}}_k^q)$ \hfill \Comment{Predicted Configuration}
            \State $\mathcal{T}^q \gets \mathcal{T}^q \cup \{\hat{X}_{k+1}^q\}$
        \EndFor
    \EndFor
    \State $q^* \gets \arg\min_q \sum_{k=1}^L \left( \min_j \Delta (Y_j, \hat{X}_k^q) \right)$ \hfill \Comment{Optimal}
    \State $\leftidx{^b}{u}_t \gets \leftidx{^b}{\hat{u}}_0^{q^*}$ \hfill \Comment{First Predicted Control in $\mathcal{T}^{q^*}$}
    \State \Return $\leftidx{^b}{u}_t$ 
    \end{algorithmic}
\end{algorithm}
\vspace{-20pt}
\end{figure}

\begin{figure}[h]
\centering
\includegraphics[width=0.75\columnwidth]{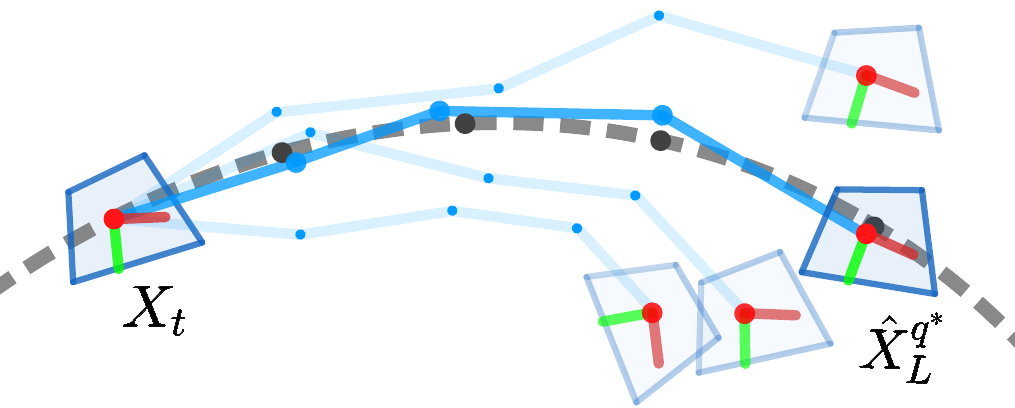}
\caption{Trajectory simulation and optimization by MPC.
By iteratively propagating the object's configuration with the estimated models $\Gamma$ and $\Gamma^{-1}$ and random perturbations, a bunch of trajectories (blue) are simulated to a horizon $L$.
The optimal one (thick blue), which is closest to the reference trajectory (dashed), is selected to extract the control input for execution.}
\label{fig:mpc_traj}

\end{figure}

As illustrated in Fig.~\ref{fig:mpc_traj}, with the object currently at configuration $X_t$, the approximated models $\Gamma$ and $\Gamma^{-1}$ are used to simulate $Q$ controlled trajectories of the object's configuration, up to a prediction horizon $L$.
Each trajectory, $\mathcal{T}^q = \{\hat{X}_k^q\}_{k=0}^L$, $q=1, \cdots, Q$, is simulated by iteratively propagating the system through Eq.~\eqref{eq:prop1}-\eqref{eq:prop4} with $\hat{X}_0^q = X_t$.

At each iteration of trajectory simulation,
assuming the object configuration is $\hat{X}_k^q$ as predicted at the current step, the nearest waypoint in the reference trajectory $Y_{j^*} \in \mathcal{Y}$ is found by Eq.~\eqref{eq:prop1}.
The next waypoint 
$Y_{j^*+1}$
is used as a reference goal to calculate the object's desired motion $\leftidx{^b}{\hat{g}}_k^q$ by Eq.~\eqref{eq:prop2}.
We perturb this desired motion by applying a random transformation $g_\xi \in SE(2)$ to it, which is generated by randomly sampling a small rotation and translation. 
This perturbed motion is passed to the inverse model $\Gamma^{-1}$ for predicting a desired control $\leftidx{^b}{\hat{u}}_k^q$ by Eq.~\eqref{eq:prop3}.
Assuming this predicted control $\leftidx{^b}{\hat{u}}_k^q$ is executed next, we can forward propagate the system by $\Gamma$ to predict the outcome configuration $\hat{X}^q_{t+1}$ as in Eq.~\eqref{eq:prop4}, to be used in the next iteration.
It is worth noting that the $Q$ trajectories are simulated differently due to the randomness in the perturbation $g_\xi$.
\begin{alignat}{4} 
    \label{eq:prop1} & j^* = \underset{j \in \{1, \cdots, M\}}{\arg\min} \Delta (\hat{X}_k^q, Y_j)\\
    \label{eq:prop2} & \leftidx{^b}{\hat{g}}_k^q = \hat{X^q_k}^{-1} \cdot Y_{j^*+1}\\
    \label{eq:prop3} & \leftidx{^b}{\hat{u}}_{k}^q = \Gamma^{-1}(\leftidx{^b}{\hat{g}}_k^q \cdot g_\xi)\\
    \label{eq:prop4} & \hat{X}_{k+1}^q = \hat{X}_k^q \cdot \Gamma(\leftidx{^b}{\hat{u}}_k^q)
\end{alignat}

Over the $Q$ differently simulated trajectories, the optimal trajectory $\mathcal{T}^{q^*}$ with the lowest cost will be found by Eq.~\eqref{eq:optimal_traj}, and 
 the first predicted control in $\mathcal{T}^{q^*}$ will be extracted for execution by the robot. 
As the entire procedure of trajectory simulation and optimization (visualized in Fig.~\ref{fig:mpc_traj}) is performed at each time step by MPC, our unified framework is able to enable the robot to precisely manipulate the object through generated pushing actions in a closed loop.
\begin{equation}
\label{eq:optimal_traj}
    q^* = \underset{q \in \{1, \cdots, Q\}}{\arg\min} \sum_{k=1}^L \left( \min_{j \in \{1, \cdots, M\}}\Delta (Y_j, \hat{X}_k^q) \right)
\end{equation}

The generated control approaches optimal control as the number of simulated trajectories $Q$ increases.
When $Q = 0$, our control scheme will be reduced
to executing a control directly predicted by the inverse model $\Gamma^{-1}$, thus becoming greedy and non-optimal.

\section{Smoothening the Execution of Controls}
\label{sec:exec}
By our representation of controls defined in Sec.~\ref{sec:rep} and shown in Fig.~\ref{fig:re},
whenever executing a control generated by MPC, the robot needs to first retreat its gripper back to a point $P$ on the virtual circle before approaching the object to push it.
This makes the gripper move back and forth, causing the object not continuously pushed by the robot.
To this end, we implemented an optimized execution in Alg.~\ref{alg:exec} by smoothly connecting adjacent control executions to facilitate continuous manipulation of the object.

\rev{We first calculate the Euclidean distance between the current control $\leftidx{^b}{u}_t = (\alpha_t, \beta_t)$ and the control executed at the previous time step $\leftidx{^b}{u}_{t-1} = (\alpha_{t-1}, \beta_{t-1})$.} When the \rev{distance} is less than a threshold $\sigma$, instead of reaching the point $P$, 
the new execution strategy will let the gripper first approach a point $P'$ closer to the object than $P$ and then move the gripper in the desired direction to apply the push.
The point $P'$ is chosen such that 1) its distance to the object should equal the current distance between the gripper and the object (i.e., $r$ at line 3 in Alg.~\ref{alg:exec}); 2) the direction of the line segment $\overline{PP'}$ matches the desired pushing direction of $\leftidx{^b}{u}_t$.
To guarantee $\overline{PP'}$ aligns with the desired pushing direction, we solve an auxiliary angle $\gamma_t$ at line 5 in Alg.~\ref{alg:exec} by the Law of Sines in a triangle.

\begin{figure}
\begin{algorithm}[H]
    \caption{SmoothenedExecute($\cdot$)}
    \label{alg:exec}
    \footnotesize
    \begin{algorithmic}[1]
    \Require Object's current configuration $X_t$, control $\leftidx{^b}{u}_t$
    \Ensure Object's outcome configuration $X_{t+1}$
        \State $\leftidx{^b}{u}_{t-1} \gets \Call{GetLastControl}$ \hfill \Comment{From the Previous Time Step}
        \If{$\lVert \leftidx{^b}{u}_{t-1} -  \leftidx{^b}{u}_t \rVert < \sigma $}
            \State $p_t \gets \Call{GetObjectPosition}{X_t}$ \hfill \Comment{Object's Position}
            \State $p_{EE} \gets \Call{GetHandPosition}$ \hfill \Comment{End-Effector's Position}
            \State $r \gets \lVert p_t - p_{EE} \rVert$
            \State $(\alpha_t, \beta_t) \gets \leftidx{^b}{u}_t$
            \State $\gamma_t \gets \alpha_t  + \beta_t - \arcsin{\frac{R\sin{\beta_t}}{r }}$ \hfill \Comment{$R:$ The Virtual Circle Radius}
            \State $P' \gets (r\cos{\gamma_t}, r\sin{\gamma_t})$  \hfill \Comment{In Object's Body Frame}
            \State $\Call{MoveHandTo}{P'}$
            \State $X_{t+1} \gets \Call{Push}{\overline{PP'}, d}$ \hfill \Comment{$\overline{PP'}$: Direction; $d$: Distance}
        \Else
            \State $X_{t+1} \gets \Call{Execute}{\leftidx{^b}{u}_t}$ \hfill \Comment{Not Smoothened}
        \EndIf
        \State \Return $X_{t+1}$
    \end{algorithmic}
\end{algorithm}
\vspace{-20pt}
\end{figure}
\section{Experiments}
\label{sec:experiments}
We evaluated our proposed framework UNO Push on a Franka Emika Panda robot, with the manipulated object tracked by cameras via AprilTags~\cite{olson2011apriltag}.
Through real-world experiments, we would like to investigate the performance of our framework from two aspects: 1) without any explicit modeling of the system or physical properties like object shapes and inertial parameters, what precision can be achieved by the proposed UNO Push;
and 2) how the robustness of the approximated system models is affected by different settings of the framework \rev{and external perturbations}.
Moreover, in comparison with a state-of-the-art baseline \rev{which relies on system transition models trained directly on the target objects} \cite{bauza2018data}, our framework achieved comparable manipulation precision \rev{in manipulating objects with unknown shapes and mass distributions}. 

\begin{figure}[htbp]

\centering
\includegraphics[width=0.99\columnwidth]{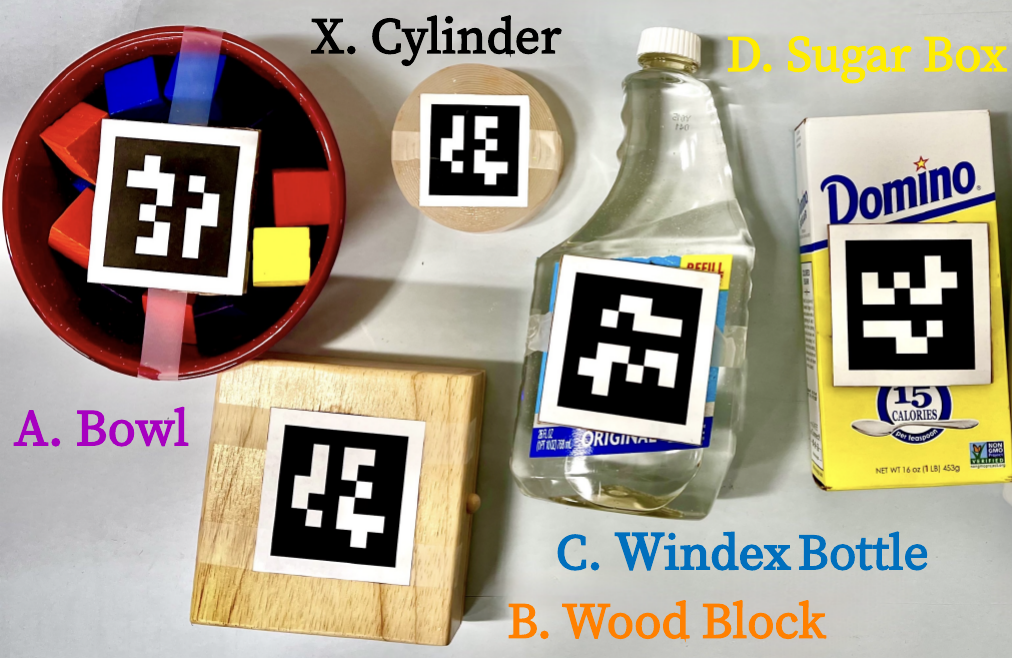}
\caption{The objects used in the experiments. A. Bowl (YCB $\#0024$) B. Wood block (YCB $\#0071$) C. Windex bottle (YCB $\#0022$) D. Sugar box (YCB $\#0004$) X. A Cylinder object}
\label{fig:objects}
\vspace{-0.2cm}
\end{figure}

We selected several objects from the YCB dataset~\cite{calli2017yale} for experiments, as shown in Fig.~\ref{fig:objects}. 

\rev{In all our experiments, we consistently ran our proposed framework (Alg. \ref{alg:uno}) at a frequency of $20$Hz.} In all the experiments except for those in Sec.~\ref{sec:exp1}, no matter which target object was manipulated for testing, the non-parametric system models were learned only by the cylinder (object X in Fig.~\ref{fig:objects}) with $10$ sampled actions.
For quantitative evaluation, we used a single circle with a radius of $0.15$m as our reference trajectory across all the experiments, as shown in Fig.~\ref{fig:first}.
To evaluate the precision of object manipulation, 
we used Mean Absolute Error (MAE) as our metric, which was computed by averaging the absolute distance between the object's actual trajectory and the reference trajectory over the entire manipulation.

\subsection{Robustness of Non-parametric Models}
\label{sec:exp1}

We tested UNO push with four different experimental settings described in Fig.~\ref{tab:settings}.
In different settings, the non-parametric system models were learned on object X or on the target object itself being manipulated (objects A-D in Fig.~\ref{fig:objects}), and the online model update was opted to be enabled or not. 
We fixed the hyperparameters of MPC to $L=20$ and $Q=50$.
Evaluated with different numbers of data points for model learning (i.e., $N = 5, 10, 20, 50$), the manipulation performance under different settings is reported in Fig.~\ref{fig:ex2}.
The reported MAE was averaged over four trials, each on a different tested object (objects A-D).

\begin{figure}[htbp]
\centering
\setlength{\tabcolsep}{6pt}
\renewcommand{\arraystretch}{1.2}
\footnotesize
\centering
\begin{tabular}{c| c}
    \hline
    Setting & Description\\
    \hline
    \hline
    \#1 & \rev{model learned} on the \textbf{cylinder X} \rev{and transferred} \\ &\rev{to the target object} \emph{without} online update\\
    \hline
    \#2 & \rev{model learned} on the \textbf{cylinder X} \rev{and transferred} \\ &\rev{to the target object} \emph{with} online update\\
    \hline
    \#3 & \rev{model learned} on the \textbf{target object} \emph{without} online update\\
    \hline
    \#4 & \rev{model learned} on the \textbf{target object} \emph{with} online update\\
    \hline
\end{tabular}
\vspace{1pt}
\caption{Four different settings for model estimation in the experiments of Sec.~\ref{sec:exp1}.}
\label{tab:settings}
\end{figure}

\begin{figure}[htbp]
\centering
\vspace{-85pt}
%
%
\definecolor{mycolor1}{rgb}{0.23529,0.72941,0.32941}%
\definecolor{mycolor2}{rgb}{0.95686,0.76078,0.05098}%
\definecolor{mycolor3}{rgb}{0.85882,0.19608,0.21176}%
\definecolor{mycolor4}{rgb}{0.28235,0.52157,0.92941}%
\definecolor{mycolor5}{rgb}{1.00000,0.54902,0.00000}%
\begin{tikzpicture}

\begin{axis}[%
width=0.85\columnwidth,
height=0.4\columnwidth,
scale only axis,
grid=both,
grid style={line width=.1pt, draw=gray!20},
major grid style={line width=.2pt,draw=gray!50},
xmin=5,
xmax=50,
xtick={5, 10, 20, 50},
xlabel={$N$: number of data points for model \rev{learning}},
ymin=3,
ymax=13,
ytick={0, 5, 10, 15},
minor y tick num=5,
ylabel={Average MAE [mm]},
title style={font=\bfseries},
legend style={legend cell align=left, align=left, legend columns = 2, fill=none, draw=none, at={(0.02,0.85)},anchor=west}
]
\addplot [color=mycolor1, line width=1.0pt, mark=o, mark options={solid, mycolor1}]
  table[row sep=crcr]{%
5	7.38728338040275\\
10	8.28065576200503\\
20	9.1247826403994\\
50	10.5247749213879\\
};
\addlegendentry{Setting \#1}

\addplot[area legend, draw=none, fill=mycolor1, fill opacity=0.3, forget plot]
table[row sep=crcr] {%
x	y\\
5	7.82800135240217\\
10	9.22275242309055\\
20	10.345042848807\\
50	12.0318679810584\\
50	9.01768186171739\\
20	7.90452243199178\\
10	7.33855910091952\\
5	6.94656540840333\\
}--cycle;

\addplot [color=mycolor2, line width=1.0pt, mark=o, mark options={solid, mycolor2}]
  table[row sep=crcr]{%
5	6.36306831241539\\
10	5.52737557331692\\
20	5.95756609555554\\
50	6.77038746061467\\
};
\addlegendentry{Setting \#2}

\addplot[area legend, draw=none, fill=mycolor2, fill opacity=0.3, forget plot]
table[row sep=crcr] {%
x	y\\
5	6.89656504426379\\
10	5.83908340802797\\
20	6.34289560366123\\
50	7.26563481182223\\
50	6.27514010940712\\
20	5.57223658744986\\
10	5.21566773860588\\
5	5.82957158056699\\
}--cycle;
\addplot [color=mycolor3, line width=1.0pt, mark=o, mark options={solid, mycolor3}]
  table[row sep=crcr]{%
5	7.15741487043947\\
10	6.88307357854357\\
20	5.87641180410048\\
50	4.22659402988049\\
};
\addlegendentry{Setting \#3}

\addplot[area legend, draw=none, fill=mycolor3, fill opacity=0.3, forget plot]
table[row sep=crcr] {%
x	y\\
5	7.43058123744771\\
10	7.38527327029879\\
20	6.19039360816855\\
50	4.69758289811642\\
50	3.75560516164456\\
20	5.5624300000324\\
10	6.38087388678835\\
5	6.88424850343124\\
}--cycle;
\addplot [color=mycolor4, line width=1.0pt, mark=o, mark options={solid, mycolor4}]
  table[row sep=crcr]{%
5	6.25685174871789\\
10	5.13462250883543\\
20	4.93704164228756\\
50	3.73195043179985\\
};
\addlegendentry{Setting \#4}

\addplot[area legend, draw=none, fill=mycolor4, fill opacity=0.3, forget plot]
table[row sep=crcr] {%
x	y\\
5	6.79868445088354\\
10	5.6864578315978\\
20	5.55471809592879\\
50	4.17497575110581\\
50	3.28892511249389\\
20	4.31936518864633\\
10	4.58278718607305\\
5	5.71501904655224\\
}--cycle;
\end{axis}

\begin{axis}[%
width=4.167in,
height=2.604in,
at={(0in,0in)},
scale only axis,
xmin=0,
xmax=1,
ymin=0,
ymax=1,
axis line style={draw=none},
ticks=none,
axis x line*=bottom,
axis y line*=left
]
\end{axis}
\end{tikzpicture}%
\caption{
Performance evaluation under different settings defined in Fig.~\ref{tab:settings}, with different number of data points for model \rev{learning and transfer}. The shaded regions indicate the standard deviations.}
\label{fig:ex2}
\vspace{-0.3cm}
\end{figure}
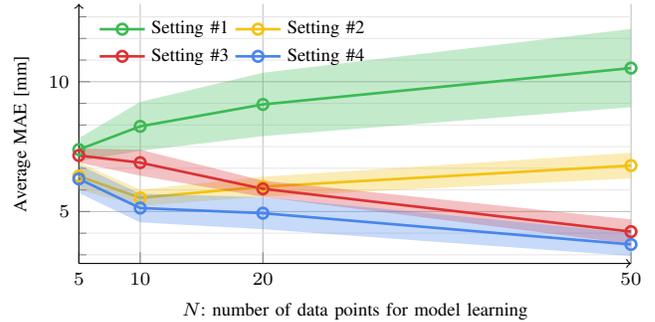

First, without online model update enabled (Setting \#1), more data collected with the cylinder X can cause worse manipulation performance on a novel object.
Trained with more data collected on the cylinder, the estimated models were more likely to be overfitted to the cylinder and thus made more inaccurate predictions about the motions of the novel object.
Second, by comparing \#1 against \#2 (or \#3 against \#4), no matter which object or how much data was used to \rev{pre-train} the model, online model udpate always facilitated better manipulation.This is because online model update enables more extensive exploration of the system models specific to the object being manipulated, resulting in more accurate model estimation. Last, when the models were learned on the target object being manipulated rather than on the cylinder (\#3 or \#4), more data points enabled better model estimation due to more valid explorations of the underlying system dynamics.

In general, the results have shown that a small amount of data is sufficient for our framework to achieve high-precision manipulation. With only $10$ actions explored to train the initial models of $\Gamma$ and $\Gamma^{-1}$ (under Setting \#2), an averaged MAE of less than $6$mm was achieved on novel objects.

\subsection{MPC Performance Evaluation}

\begin{figure}[htbp]
\centering
\vspace{-80pt}
%
%
\definecolor{mycolor1}{rgb}{0.23529,0.72941,0.32941}%
\definecolor{mycolor2}{rgb}{0.95686,0.76078,0.05098}%
\definecolor{mycolor3}{rgb}{0.85882,0.19608,0.21176}%
\definecolor{mycolor4}{rgb}{0.28235,0.52157,0.92941}%
\definecolor{mycolor5}{rgb}{1.00000,0.54902,0.00000}%
\begin{tikzpicture}

\begin{axis}[%
width=0.85\columnwidth,
height=0.4\columnwidth,
scale only axis,
grid=both,
grid style={line width=.1pt, draw=gray!20},
major grid style={line width=.2pt,draw=gray!50},
xmin=0,
xmax=200,
xtick={0, 10, 20, 50, 200},
xlabel={$Q$: number of simulated trajectories},
ymin=4,
ymax=12,
ytick={0, 5, 10, 15},
minor y tick num = 5,
ylabel={Average MAE [mm]},
title style={font=\bfseries},
legend style={legend cell align=left, align=left, draw=none, fill=none}
]
\addplot [color=mycolor1, line width=1.0pt, mark=o, mark options={solid, mycolor1}]
  table[row sep=crcr]{%
0   10.28102782473155\\
10	9.45308200404989\\
20	8.38710507941326\\
50	6.97009116384511\\
200	6.22398835103264\\
};
\addlegendentry{L = 5}

\addplot[area legend, draw=none, fill=mycolor1, fill opacity=0.3, forget plot]
table[row sep=crcr] {%
x	y\\
0   11.54490266464512\\
10	10.5541060047483\\
20	9.29797635167775\\
50	7.4325275853128\\
200	6.77584952130546\\
200	5.67212718075982\\
50	6.50765474237741\\
20	7.47623380714877\\
10	8.35205800335149\\
0   9.017152984817981\\
}--cycle;
\addplot [color=mycolor2, line width=1.0pt, mark=o, mark options={solid, mycolor2}]
  table[row sep=crcr]{%
0   10.57324658464478\\
10	7.59201769801253\\
20	6.10503527142945\\
50	5.73305950116145\\
200	5.29975258946787\\
};
\addlegendentry{L = 10}

\addplot[area legend, draw=none, fill=mycolor2, fill opacity=0.3, forget plot]
table[row sep=crcr] {%
x	y\\
0   11.85072301405311\\
10	8.11961359174643\\
20	6.60393226589431\\
50	6.06009183194605\\
200	5.71823688825335\\
200	4.88126829068238\\
50	5.40602717037685\\
20	5.60613827696459\\
10	7.06442180427863\\
0   9.29577015523645\\
}--cycle;
\addplot [color=mycolor3, line width=1.0pt, mark=o, mark options={solid, mycolor3}]
  table[row sep=crcr]{%
0   10.68401870835124\\
10	6.60144261620344\\
20	5.63605919646108\\
50	5.01939662918662\\
200	4.70972198797672\\
};
\addlegendentry{L = 20}

\addplot[area legend, draw=none, fill=mycolor3, fill opacity=0.3, forget plot]
table[row sep=crcr] {%
x	y\\
0   11.99320897343908\\
10	7.08328042889222\\
20	5.90123475806915\\
50	5.60469675397212\\
200	5.24724431047006\\
200	4.17219966548338\\
50	4.43409650440112\\
20	5.370883634853\\
10	6.11960480351466\\
0   9.3748284432634\\
}--cycle;
\end{axis}

\begin{axis}[%
width=4.167in,
height=2.604in,
at={(0in,0in)},
scale only axis,
xmin=0,
xmax=1,
ymin=0,
ymax=1,
axis line style={draw=none},
ticks=none,
axis x line*=bottom,
axis y line*=left
]
\end{axis}
\end{tikzpicture}%

\caption{Performance evaluation w.r.t. different $L$ and $Q$ in MPC, where the shaded regions indicate the standard deviations.}
\label{fig:ex1}
\vspace{-8pt}
\end{figure}
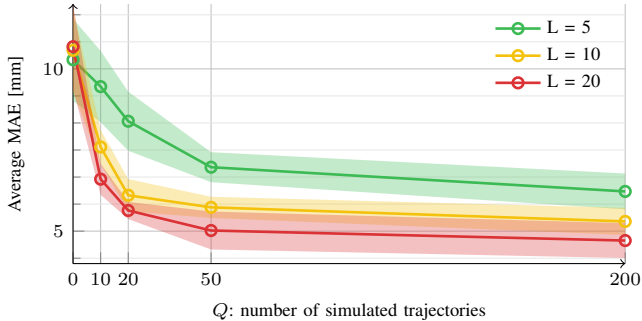

In this experiment, we evaluated the precision 
achieved with our framework by varying the prediction horizon $L$ and the number of simulated trajectories $Q$ in MPC.
These two parameters were selected from $L = 5, 10, 20$ and $Q = 0, 10, 20, 50, 200$.
For each different combination of $L$ and $Q$, we conducted four trials on four objects (Object A-D shown in Fig.~\ref{fig:objects}).
We averaged the MAE over the four trials to summarize the results in Fig.~\ref{fig:ex1}.
The system models were estimated under Setting \#2 defined in Fig.~\ref{tab:settings}.
That is, 
we saved $\Gamma$ and $\Gamma^{-1}$ learned by manipulating the cylinder X through $10$ actions, and \rev{transferred them as the system models} for all other objects while updating them online.

From the results, we can see that the manipulation precision was improved when more simulated trajectories were generated.
This is because, with more perturbed trajectories to optimize over, MPC is able to search more extensively to find a more optimal control that has a higher probability of pushing the object in its desired direction.
Moreover, with the same number of simulated trajectories, a larger prediction horizon $L$ improves the manipulation performance.
The reason is that a large $L$ enables the system models to predict the long-horizon outcome of controls, providing more reliable evidence for generating effective controls.
Particularly, when $Q=0$, MPC was directly executing the predicted control from $\Gamma^{-1}$, which in most cases was not effective enough due to the inaccuracy of model approximation.
This in turn verifies the importance of optimization in MPC.
In general, with inaccurately approximated system models, 
MPC could achieve millimeter-level precision of manipulation.

\subsection{Comparison with Baseline}

\begin{figure}[htbp]
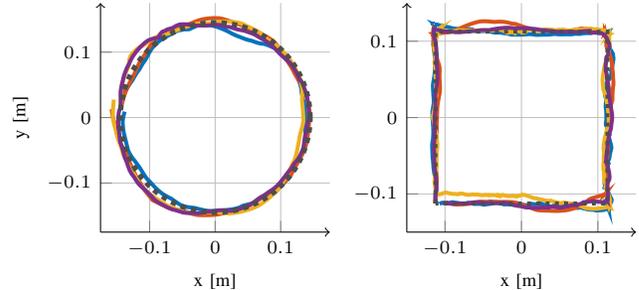

    \centering
    \begin{minipage}[b]{0.49\linewidth}
        \centering
        \input{figs/circle_traj.tex}
        
    \end{minipage}
    \begin{minipage}[b]{0.49\linewidth}
        \centering
        \input{figs/square_traj.tex}
        
    \end{minipage}
    \caption{The object's actual trajectories \revi{of UNO Push in experiments repeatd from the baseline \cite{bauza2018data}}, with the goal of tracing a circle (left) and a square (right) reference trajectory.  The lines are color-coded in the same way as the objects A-D shown in Fig.~\ref{fig:objects}.}
    \label{fig:qex1}
\vspace{-10pt}
\end{figure}

In a comparative evaluation against a baseline~\cite{bauza2018data}, as reported in Fig.~\ref{fig:exc}, \rev{our framework achieved comparable precision to the analytical method and the data-driven method in the baseline~\cite{bauza2018data} for both circle and square trajectories, while requiring only $10$ initial data points without any sophisticated modeling, significantly less than the baseline's $5,000$ data points under this specific test. \revi{As illustrated in Fig.~\ref{fig:qex1}, the outcome trajectories of the executions across various objects were consistently good, regardless of the object manipulated or the shape of the reference trajectory.} }

\rev{It's worth noting that other results from the baseline have achieved a very good accuracy of 14mm with just $10$ data points. We used the results with $5,000$ data points reported in \cite{bauza2018data} as our comparison baseline in Fig.~\ref{fig:exc}, since that was the best result in \cite{bauza2018data} with the most comprehensive evaluation for both circle and square trajectories.}

\rev{Moreover, while the baseline~\cite{bauza2018data} was trained offline for a square object of uniform mass, our approach does not require any \textit{a priori} knowledge about the contact geometries or physics, \revi{nor does it require any object-specific training}. This makes it a more generalizable and low-barrier solution for nonprehensile manipulation tasks in everyday tasks.}

\begin{figure}[htbp]
\centering
\setlength{\tabcolsep}{7pt}
\footnotesize
\centering

\begin{tabular}{c|| c | c}
    \hline
    Trajectory & Method & Error (mm)\\
    \hline
    \multirow{3}{*}{Circle} & \cite{bauza2018data} (analytical), $v= 20          \mathrm{~mm/s}$ & 2.89\\
                         & \cite{bauza2018data} (data-driven), $v= 20 \mathrm{~mm/s}$ & 6.53\\
                         & UNO Push (ours) $v= 50 \mathrm{~mm/s}$ & 5.87\\
    \hline
    \multirow{3}{*}{Square} & \cite{bauza2018data} (analytical), $v= 50          \mathrm{~mm/s}$ & 4.95\\
                         & \cite{bauza2018data} (data-driven), $v= 50          \mathrm{~mm/s}$ & 6.60\\
                         & UNO Push (ours), $v= 50          \mathrm{~mm/s}$ & 5.42\\
    \hline
\end{tabular}
\vspace{1pt}
\caption{Performance evaluation by comparison with a baseline, where $v$ denotes the motion velocity of the robot gripper.}
\label{fig:exc}
\vspace{-5pt}
\end{figure}

\subsection{Qualitative Evaluation}

\begin{figure}[htbp]
\centering
\includegraphics[width=0.99\columnwidth]{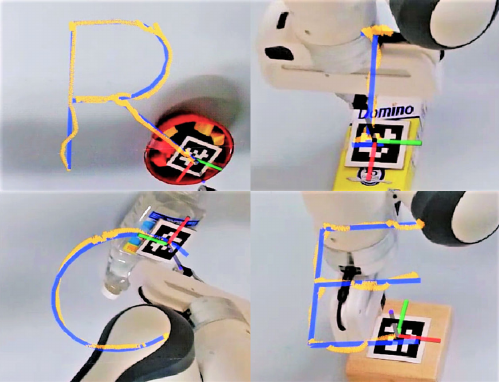}

\caption{
Example object pushing manipulation to trace letter-shaped trajectories with four different objects. The blue lines are the reference trajectories, and the yellow lines are the actual trajectories of the object.}
\label{fig:RICE}
\vspace{-10pt}
\end{figure}

\revi{To further test the robustness of UNO Push, three additional evaluation tasks were performed.} First, we applied our framework to trace letter-shaped trajectories of ``R'', ``I'', ``C'', and ``E'', with different objects, as shown in Fig.~\ref{fig:RICE}. \revi{The results show the robustness of our framework to manipulate through complex trajectories even with sharp turns (from $90^\circ$ to $180^\circ$).}

\begin{figure}[htbp]
\centering
\includegraphics[width=1\columnwidth]{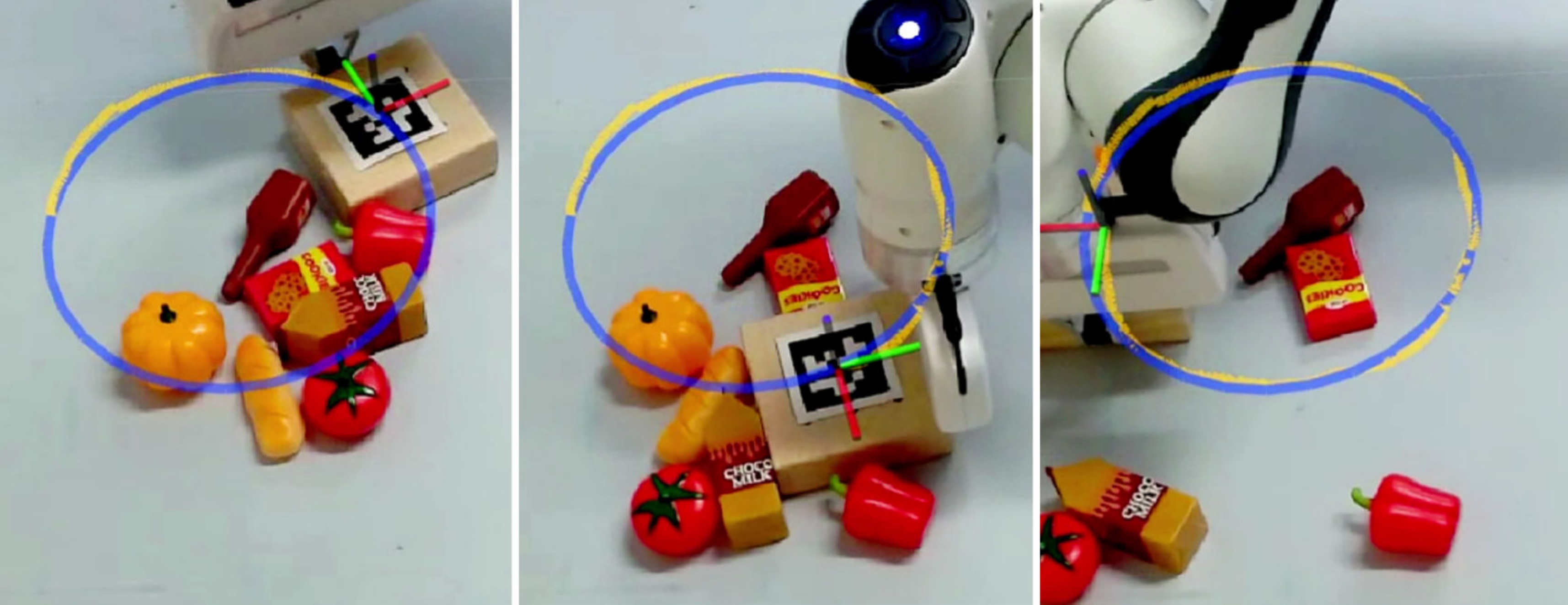}

\caption{Example object pushing manipulation through a cluttered area \rev{to demonstrate the performance under unmodeled external  perturbations}.}
\label{fig:obstacle}
\vspace{-10pt}
\end{figure}

As demonstrated in Fig.~\ref{fig:obstacle}, our framework could also push objects through unknown perturbations caused by interactions with the object clutter\rev{, as assisted by the sufficient frequency of online model updates and MPC ($20$Hz)}. Finally, as shown in Fig.~\ref{fig:first}, even with an unmodeled object grasped by the robot gripper, our method was still able to push the target object through object-object contact.
This again demonstrates that our framework is able to work with different physical uncertainties and unmodeled contacts.

\section{Conclusion}



In this paper, we proposed a unified framework, named UNO Push, for pushing-based nonprehensile object manipulation.
It unifies system model estimation, action generation, and control through light-weight non-parametric learning and closed-loop MPC.
With extensive experiments on a real 7-DoF robot, we showed that our framework can achieve millimeter-level manipulation precision, without requiring heavy data collection, sophisticated system modeling\rev{, or offline training on the target object}. 

\rev{Our work offers broad possibilities for applications of pushing-related manipulation tasks. As UNO Push provides an efficient approach for fundamental pushing tasks like trajectory tracking, it could serve as a low-level controller or motion primitive for various complex nonprehensile manipulation tasks. These tasks, such as non-prehensile rearrangement and multi-modal manipulation planning in household environments, often require precise manipulation of different objects without accurate object models and object-specific training.  }

\rev{Despite promising results, our method has limitations. First, our approach is derived under the assumption of planar pushing and quasi-static scenarios, so it may be difficult to deal with more dynamic motions of the object like rolling and flipping, or more dynamic actions like hitting and throwing objects on a 2D plane. Second, the target object is assumed to be a rigid body, and the proposed approach might not directly perform effectively on deformable or soft objects.}

In future work, \rev{we plan to extend the framework to tasks that involve more complex physical interactions, such as pushing a group of multiple objects together under formation constraints, as well as pushing objects that move with motions beyond quasi-static patterns, e.g., rolling. We are also interested in exploring the possibilities of applying the UNO Push to more complex nonprehensile rearrangement tasks, such as multi-object sorting. These tasks would require multi-modal manipulation skills, such as grasping and pushing for task and motion planning.}





\bibliographystyle{IEEEtran}
\bibliography{reference}

\end{document}